# NEW TECHNOLOGIES FOR SUSTAINABLE URBAN TRANSPORTATION IN EUROPE

Michel Parent

INRIA - IMARA

France

michel.parent@inria.fr

**ABSTRACT**

In the past few years, the European Commission has financed several projects to examine how new technologies could improve the sustainability of European cities. These technologies concern new public transportation modes such as guided buses to form high capacity networks similar to light rail but at a lower cost and better flexibility, PRT (Personal Rapid Transit) and cybercars (small urban vehicles with fully automatic driving capabilities to be used in carsharing mode, mostly as a complement to mass transport). They also concern private vehicles with technologies which could improve the efficiency of the vehicles as well as their safety (Intelligent Speed Adaptation, Adaptive Cruise .Control, Stop&Go, Lane Keeping,…) and how these new vehicles can complement mass transport in the form of car-sharing services.

**INTRODUCTION**

Cities throughout the world are facing numerous challenges in this new century. They are moving towards increased concentration into megalopolises with more than 100 cities expected to be over 10 millions inhabitants during the next fifty years while many smaller ones might face a bleak future if they are not "connected" into them.

The concentration of large populations into these megalopolises has been made possible by the development of two forms of transportation. The first one has been developed in the 19$^{th}$ century with the trains and subways. The second one has been issued from the development of the automobile with two phases. The first phase occurred in the first half of the 20$^{th}$ century with the development of buses and taxis giving the citizen a new flexibility in their transport not possible with the subways and trains, in particular for reaching peripheral locations. The second phase occurred in the second half of the 20$^{th}$ century with the democratisation of the private automobile and the possibility for every citizen to live outside the city, in what became the suburbs.

The consequences of all these innovations in transports have been enormous in terms of urbanism with the freedom for the population to choose where to live, to work and to develop other activities. However, it has also lead to problems concerning safety and quality of life of the citizens with pollution, noise, excessive travel times, and more and more difficulties to move around without access to an automobile. Another problem concern the excessive energy cost and the massive utilisation of fossil fuel vehicles with problems linked to oil dependency and global warming.

Several recent European Projects have been funded to face these challenges. Numerous solutions have been proposed and they can be divided into two approaches:

- a regulation of the demand for transport through a better use of the land (land-use approach) with projects some of which have been grouped into the LUTR cluster (http://www.lutr.net) ,
- new forms of urban transport which are more sustainable and offer a better mobility to the whole of the population with also a number of projects grouped into the NetMobil cluster (http://www.netmobil.org).

This paper will present the results of these two European approaches and the technologies that can be relied upon to improve mobility while minimising the problems. We believe that these approaches could also be very beneficial in other parts of the world although the technologies to be applied could be somewhat different due to different business cases.

**LAND USE AND URBANISM**

The LUTR cluster has identified problems and has issued some recommendations which are important for the future of sustainable cities:

*Across Europe there is a common challenge to improve the quality of life in urban communities, and to ensure the competitiveness of cities, whilst promoting sustainable development. All cities face common challenges relating to air quality, noise, urban sprawl, traffic congestion, waste, economic competitiveness, job creation, security, social inclusion, and maintaining the built environment, cultural heritage, and a deteriorating infrastructure.*

And through annual State of the Art Reviews (SoARs), LUTR researchers have brought to light the following directions for moving into more sustainable cities:

*These passenger and freight transport trends and projections point to the following land use and transport policy instruments and processes being in place by 2030:*

- *Travel Demand Management,*
- *Parking controls,*
- *Road user charging for passenger and freight vehicles,*
- *Public transport development,*
- *Innovative modes,*
- *Home delivery and services,*
- *Air transport developments,*
- *Freight transport regulations,*
- *Rail freight,*
- *Urban distribution centres,*
- *New technology,*
- *Properly integrated land use and transport planning,*

- *Land use planning that favours urban regeneration and polycentric development,*
- *More harmonised land use and transport strategy development, forecasting, appraisal and implementation to promote sustainability and quality of life,*
- *Targets and indicators to support the above,*
- *Public participation.*

*From the perspective of citizens, all this means that many will live in densely populated urban areas at various points throughout their lives, notably, between leaving home and start families, and again in older age. Some families will also continue to live in these urban areas where polycentric development creates family friendly regeneration, i.e., including local schools, play groups, leisure destinations catering for children etc, as well as creating urban villages where people know each other, creating a safe environment for children to grow up in. Such environments also provide a better quality of life for everybody, especially single person households (be they young or old), those with mobility impairments, and those on a low income. Urbanites will have shorter everyday travel distances, many of which will be undertaken by public transport (including innovative modes), on foot or, where terrain and weather permits, by bicycle.*

The propositions clearly call for a strong reduction of private car use through a better structure of the city, with a polycentric organisation for the largest ones, through better use of soft modes such as cycling or walking, and through better public transport, in particular with new technologies.

**NEW TECHNOLOGIES**

It was the focus of the NetMobil Cluster to examine how new technologies could improve the sustainability of European cities. These technologies concern new public modes such as guided buses to form high capacity networks similar to light rail but at a lower cost and better flexibility, PRT (Personal Rapid Transit) and cybercars (small urban vehicles with fully automatic driving capabilities to be used in carsharing mode, mostly as a complement to mass transport). They also concern private vehicles with technologies which could improve the efficiency of the vehicles as well as their safety (Intelligent Speed Adaptation, Adaptive Cruise .Control, Stop&Go, Lane Keeping,…).

The conclusions of this Cluster can be found in their final document:

*Recommendations for the implementation of potentially sustainable personal urban transportation systems through 2 approaches have been described. The first exploits developments of ADAS systems in car share schemes and private fleets to win the benefits of cleaner, greener, safer more efficient vehicles as they become available from the vehicle manufacturers. The second promotes PRT/CTS systems which similarly provide cleaner, greener, safer and more efficient transport but through a public transport approach. Both provide alternatives to the use of private cars in urban areas through exploiting automatic vehicle technologies. Both appear to offer cost effective solutions to sustainable urban transportation, and offer different and complementary solutions. Both approaches will lead ultimately to fully automatic vehicles (or dual-mode vehicles), but progress will depend on a range of factors including user acceptance, risks, legal and institutional aspects, and social and market forces.*

These conclusions are also supported by the World Business Council on Sustainability (lead by the automotive and oil industries) in their document "Mobility 2030 : Meeting the Challenge to Sustainability":

*The next 50 years may see the emergence of entirely new transport solutions. These would offer either a completely new mode of transport or would make use of a new combination of existing transport modes. New transport solutions become possible when mobility demand, in combination with support from government, the availability of the required technology and economic benefits for all stakeholders – make such solutions more attractive than those that exist.*

*Entirely new transport solutions do not appear overnight. To become available after 2030, development work would need to begin almost immediately. Numerous issues have to be addressed in advance, public acceptance obtained, and pilot projects organized. Meantime*

*developed and developing world stakeholders would likely place differing requirements in such areas as cost, infrastructure, reliability, geographical application, and logistics.*

*So-called "Cybernetic Transport Systems" (CTS) composed of road vehicles with fully automated driving capabilities are one new possibility. A fleet of such vehicles would form a transportation system for passengers or goods on a network of roads with on-demand and door-to-door capability. Cars would be under control of a central management system in order to meet particular demands in a particular environment. The size of the vehicle could vary from 1-20 seats, depending on the application. This concept is similar in many respects to another concept known as PRT (Personal Rapid Transit). But CTS offer the advantage of being able to run on normal road infrastructure. This makes them cheaper and more flexible. Existing technologies allow a relatively inexpensive "grid" to be placed over a geographic area to be served. Software drives the routing and management of the fleet of vehicles.*

*The potential of systems like CTS is great. In effect it is a high-quality public transport service that offers on-demand, door-to-door service. Moreover, the most expensive component of public transport, the driver, has been substituted. If vehicles turn out to be clean and silent, the implementation of such systems in urban areas would simultaneously reduce pollutants, noise, and congestion, improving the livability of the city. CTS also offers real mobility solutions to those who cannot drive or do not own their own vehicles. The elderly and disabled, in particular, would become mobile.*

European projects such as CyberCars ([www.cybercars.org](www.cybercars.org)), CyberMove ([www.cybermove.org](www.cybermove.org)) and EDICT (www.atsltd.co.uk) which took place between 2001 and 2005 have paved the way for the development and dissemination of such innovative urban transport approaches but further larger experiments are now expected. The next large project to be funded by the European Commission should address this issue.

New technologies and new organisational frameworks can also be used to improve the way goods are moved in the cities. The European Project ELCIDIS (www.elcidis.org) has experimented with a new form of urban delivery system using a fleet of electric vehicles. Here is the conclusion in the final report of this project which has been completed in 2002 :

*The ELCIDIS project has tested a better solution for urban logistics by approaching the subject in a dual way, taking into account the interests of all parties involved, in order to set an example for clean and efficient urban distribution in the 21st century.*

■ *By organising urban distribution using quiet and clean (hybrid) electric vehicles, the nuisance caused by distribution activities will be decreased. The improved living climate of the city will benefit residents and shoppers as well as shopkeepers.*

■ *A more efficient organisation of urban logistics is achieved by more efficient routing of the vehicles and the use of urban distribution centres (UDC). This will decrease the number of journeys made by heavy vehicles and increase traffic fluidity in urban areas. The improved accessibility of the city will benefit transport companies, shopkeepers and businesses operating in the city.*

**CONCLUSIONS**

While, the problems of mobility in European cities have been clearly identified, the solutions to be put in place are still at their infancy. It is clear that a mix of land-use policies and a shift from the private automobile to a multi-modal approach is the preferred trend. The solution for implementing the multi-modal approach (including and encouraging soft mode) goes certainly through a complementarity between high speed scheduled mass transport and individualised on demand short distance transport. However, these individual on-demand trips should not use the private automobile, in particular in the densest parts of the cities where it is not well adapted in terms of space, energy, safety, …. This is why we have now to test new solutions based on advanced city vehicles in car-sharing mode, on fully automated vehicles which run of new infrastructures (personal rapid transit or PRT), and on cross-over vehicles such as the cybercars which can run manually in mixed traffic or automatically on reserved areas (or new infrastructures).

Similar technologies can be adapted to freight transport in cities with a multimodal approach using dispatch centres outside the cities with clean vehicles running inside (in manual or automatic modes).

Such an approach would definitely be beneficial in newly developed countries such as China where the proliferation of private vehicles could lead rapidly to severe problems concerning safety, health, global warming and an overall reduction in mobility.

**REFERENCES**

Mobility 2030 : Meeting the Challenge to Sustainability. World Business Council on Sustanability. 2004.European Road Transport 2020: a Vision and Strategic Research Agenda. European Road Transport Research Advisory Council. 2004

Green Paper on Energy Efficiency or Doing More with Less. CEC, June 2005.

White Paper. European Transport Policy for 2010 : time to decide. European Commission 2001.

EPC WORKING PAPER N° 16. 12 Prescriptions for a European Sustainable Mobility Policy. The EPC Task Force on Transport. 2005

ELCIDIS, Electric Vehicle City Distribution. Final Report. TR 0048/97. European Commission, 2002.

Towards a thematic strategy on the urban environment. Communication from the EC to the Parliament. 2004

Parent Michel, Texier Pierre-Yves. "A Public Transport System Based on Light Electric Cars". Fourth International Conference on Automated People Movers. Irving, USA. March 1993.

Parent Michel, Blosseville Jean-Marc. "Automated Vehicles in Cities : A First Step Towards the Automated Highway". SAE Future Transportation Technology Conference. Costa Mesa, USA. August 11-13, 1998.

Parent Michel, Gallais Georges. "CyberCars : Review of First Projects". Ninth International Conference on Automated People Movers. Singapore. Sept.2003.

Yang M., and Parent M. "Cybernetic Technologies For Cars In Chinese Cities", *Proceeding of CityTrans China 2004, Shanghai*, China, Nov. 17-18, 2004.

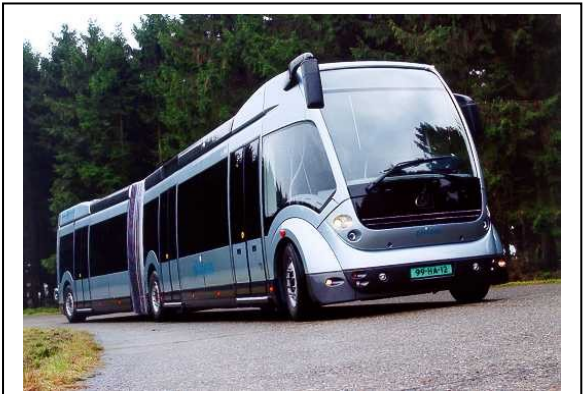

Fig. 1 : Phileas automated bus

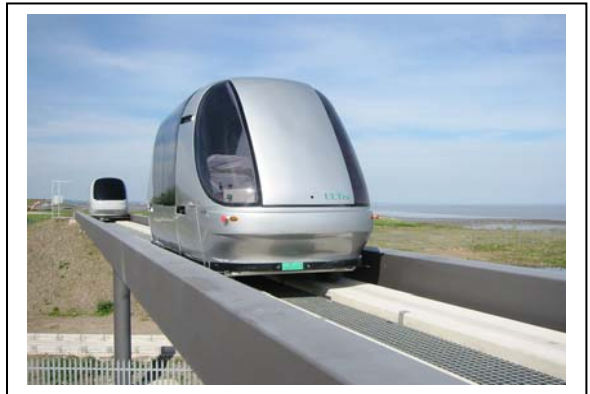

Fig. 2 : ULTra PRT

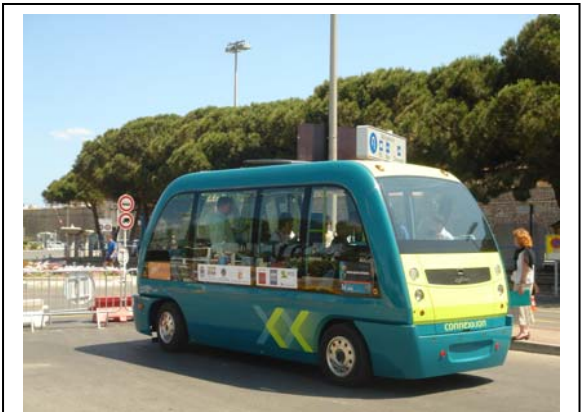

Fig. 3 : ParkShuttle cybercar

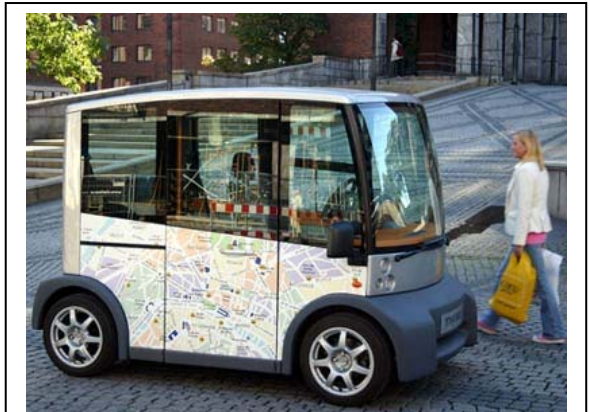

Fig. 4 : Think advanced city car